\documentclass[10pt, a4paper]{article}

\usepackage{lrec-coling2024} 
\usepackage{todonotes}

\usepackage{arabtex}
\usepackage{utf8}
\setcode{utf8}
\usepackage{xspace}
\usepackage{multirow}
\usepackage{makecell}

\newcommand{\qabas}{\cal{\textit{Qabas}}\xspace}

\newcommand{\Ar}[1]{{\scriptsize\<#1>}}

\newcommand{\TrAr}[1]{{\arabtrue\transfalse{\scriptsize\Ar{#1}}/\arabfalse\transtrue\RL{#1}\arabtrue\transfalse}\kern-0.4ex}

\title{\qabas: An Open-Source Arabic Lexicographic Database}

\abstract{
We present \qabas, a novel open-source Arabic lexicon designed for NLP applications. The novelty of \qabas lies in its synthesis of $110$ lexicons. Specifically, \qabas lexical entries (lemmas) are assembled by linking lemmas from  $110$ lexicons. Furthermore, \qabas lemmas are also linked to $12$ morphologically annotated corpora (about $2M$ tokens), making it the first Arabic lexicon to be linked to lexicons and corpora. \qabas was developed semi-automatically, utilizing a mapping framework and a web-based tool. Compared with other lexicons, \qabas stands as the most extensive Arabic lexicon, encompassing about $58K$ lemmas ($45K$ nominal lemmas, $12.5K$ verbal lemmas, and $473$ functional-word lemmas). \qabas is open-source and accessible online at \href{https://sina.birzeit.edu/qabas}{https://sina.birzeit.edu/qabas}
}

\begin{document}

\maketitleabstract

\section{Introduction}
\label{sec:introduction}
As the need for lexicographic databases in modern applications continues to grow, lexicography has evolved into a multidisciplinary field intersecting with natural language processing (NLP), ontology engineering, e-health, and knowledge management. 
Lexicons have evolved from being primarily hard-copy resources for human use to having substantial significance in NLP applications \citep{maks2009standardising, JAM19, mccrae2016open}. 



Although Arabic is a highly resourced language in terms of traditional lexicons, less attention is given to developing AI-oriented lexicographic databases. Recent efforts at Birzeit University have been devoted to digitizing traditional lexicons and publishing them online through a lexicographic search engine \cite{JA19,ADJ19}, but none of the lexicons are open-source due to copyright restrictions imposed by their owners \cite{J20}. The LDC's SAMA database \cite{SAMA}, is an Arabic lexicographic database, but it is also restricted to LDC members only.
SAMA, an extension of BAMA \cite{buckwalter2004buckwalter}, is a stem database, designed only for morphological modeling. It contains stems and their lemmas and compatible affixes.

This article proposes \qabas, a novel open-source Arabic lexicon designed for NLP applications. The novelty of \qabas lies in its synthesis of many lexical resources. Each lexical entry (i.e., lemma) in \qabas is linked with equivalent lemmas in $110$ lexicons, and with $12$ morphologically-annotated corpora (about $2M$ tokens). This linking was done through $256K$ mappings correspondences (as shown in Table \ref{tab:mapping_relations}). That is, the philosophy of \qabas is to construct a large lexicographic data graph by linking existing Arabic lexicons and annotated corpora. This enables the integration and reuse of these resources for NLP tasks. For example, by linking the lemma (\TrAr{كَرِيم2}) in SAMA with (\TrAr{كَرِيم}) in the Modern lexicon, one would integrate the morph features (stems and affixes) found in SAMA with the $4$ senses (i.e., glosses) of this lemma found in the Modern. Assuming this lemma is also linked with its $41$ word forms in the Arabic Treebank corpus \cite{Penn}, then one would compute the corpus statistics for this lemma.

\qabas was developed semi-automatically over two years, utilizing an automatic mapping framework and a web-based tool. Compared with other lexicons, \qabas is the most extensive Arabic lexicon and the first to be linked with such lexicons and corpora. The main contributions of this paper are:
\begin{itemize}
\item \textbf{Novel and open-source Arabic Lexicon} ($58K$ lemmas) linked with many NLP resources. 
\item Mappings: $256$ mapping correspondences between $110$ lexicons ($255.5K$ lemmas) and $12$ corpora ($2M$ tokens). As such, \textbf{\qabas is an Arabic lexicographic graph}, interlinking Arabic lexicons and corpora at the lemmas level.
\end{itemize} 

The paper is structured as follows: Section \ref{sec:related-work} overviews the related work, Section \ref{sec:methodology} presents the methodology, and Section \ref{sec:lemma-linking} presents lemma mapping. In Section \ref{sec:evaluation} we evaluate the coverage; and in Section \ref{sec:conclusion} we summarize our conclusions.

\section{Related Work}
\label{sec:related-work}

In recent years, many standardization efforts have been proposed for representing, publishing, and linking linguistic resources. For example, the W3C’s Lemon RDF model \citep{cimianolexicon} enables employing lexicons in ontologies and various NLP applications. Moreover, the Linguistic Linked Open Data Cloud (LLOD) \citep{mccrae2016open} used Lemon to interlink the lexical entries of several linguistic resources. The ISO's Lexical Markup Framework (LMF) standard aims at representing lexicons in a machine-readable format \citep{LMF}. 


Different encyclopedic dictionaries integrated WordNets with other resources, such as BabelNet \citep{navigli2012babelnet} and ConceptNet 5.5 \citep{speer2017conceptnet}. Compared with our work, we provide an interlinking of many lexicons and corpora, forming a lexicographic, rather than an encyclopedic graph.

Given that digitized and available Arabic lexicons are limited, there are several attempts to digitize and represent them in the standard formats. The first attempt to represent Arabic lexicons in ISO LMF standard can be found in \citep{salmon2005proposals,maks2009standardising,khemakhem2016iso}. Other attempts suggested using the W3C Lemon RDF model \cite{khalfi2016classical,JAM19}. While several online portals for Arabic lexicographic search exist (e.g., lisaan.net, almaany.com, almougem.com), each portal contains a limited number of lexicons, and their content is partially structured (i.e., available in flat text). \qabas is developed as a synthesis of $110$ lexicons that we digitized earlier \citep{JA19}.



\section{Methodology}
\label{sec:methodology}

\subsection{Scope and Objectives}
The objective of \qabas is to link existing Arabic lexicons and corpora and enable them to be integrated and re-used in NLP tasks \citep{DH21}. In other words, \qabas lemmas are used as a proxy to link between different resources, forming a large Arabic lexicographic data graph. Thus, all \qabas lemmas are collected mainly from these resources (Section \ref{sec:data_sources}). 
As such, \qabas is designed to be an open-source and open-ended project, targeting all forms of Arabic: Classical Arabic, Modern Standard Arabic, Arabic dialects, and foreign words that are transliterated and commonly used in Arabic. 

In this paper, we focus on including the morphological features for each lemma, such as the spelling(s) of the lemma, its root(s), POS, gender, number, person, and voice. Including semantic information (e.g., glosses, synonyms, relations, and translations) is not discussed in this article due to space limitations. Nevertheless, it is worth noting that based on \qabas mappings, (i) we developed a synonym extraction tool\footnote{\href{https://sina.birzeit.edu/synonyms}{https://sina.birzeit.edu/synonyms}. It can be also used to evaluate synonyms \citep{KAEMKRTM23}.}  \citep{GJJB23}; (ii) we extracted glosses and contexts from these mapped lexicons to build a large set of context-gloss pairs for Word-Sense Disambiguation \citep{HJ21b, MJK23}; and (iii) a graph representing morpho-semantic relationships in Arabic was extracted based on Arabic derivational morphology, see Figure 4 in \citep{J21}.

\subsection{Data Sources}
\label{sec:data_sources}
Among the $150$ lexicons that we previously digitized \cite{JA19}, $110$ lexicons and $12$ morphologically annotated corpora were prepared to be linked and to construct \qabas. See our copyright notice in section \ref{sec:ethical_considerations} regarding the collected resources and the sharing of \qabas.

Table \ref{tab:lexicons} categorizes the $110$ lexicons into: the LDC's SAMA  \cite{SAMA}, 
Modern lexicon \citep{omar2008contemporary}, 
Ghani lexicon \citep{abul2014ghani}, 
the Al-Waseet lexicon 
\citep{Waseet}, 
the Al-Waseet Madrasi lexicon,
the Arabic Ontology and two lexicons \cite{J21,J11}, 
the Arabic WordNet \cite{black2006introducing}, 
$40$ of the \citet{ALECSO}'s Unified dictionaries. 
We also collected $16$ lexicons produced by the Arabic Language Academies in Cairo and in Damascus \cite{Cairo,Damascus}, the Arabic Wikdata, in addition to $7$ thesauri and $37$ Other lexicons.

As we are concerned with linking the lexical entries (i.e., lemmas) in these resources, each distinct lemma is given a unique identifier. In addition, we are only concerned with linking single-word lemmas, thus multi-word lemmas are ignored at this phase, such as (\Ar{ثاني أكسيد الكربون، سرعة الضوء}). The total number of single-word lemmas in the $110$ lexicons is about $297K$ lemmas, about $255K$ ($84\%$) of which are mapped (See Table \ref{tab:lexicons}).

As shown in Table \ref{tab:corpora}, we collected $12$ Arabic corpora, especially those that are annotated with morphological features: the MSA LDC's Arabic Treebank \cite{Penn}, the MSA SALMA corpus \cite{JMHK23}, 
the Quran corpus \cite{QuranCorpus},
the Palestinian Curras and the Lebanese Baladi corpora \cite{EJHZ22}, 
the Lisan (Iraqi, Lybian, Sudanese, and Yemeni) corpora \cite{JZHNW23}, The Emirati Gummar corpus \cite{khalifa-etal-2018-morphologically},  
the Syrian Nabra corpus\cite{ANMFTM23}, and the LDC's Egyptian Treebank \cite{EgyTreeBank}.
These corpora compass $2.4M$ tokens annotated with about $144.5K$ lemmas, $84\%$ of which are mapped with \qabas; i.e., \qabas is linked with about $2M$ tokens.

\begin{table}
    \centering
        \scriptsize 
    \begin{tabular}{|l|r|r|r|} \hline 
        \textbf{Lexicon}&\makecell[c]{\textbf{Unique} \\\textbf{Lemmas}}& \makecell[c]{\textbf{Lemmas} \\\textbf{mapped}}\\ \hline

SAMA&$40,639$&$40,330^{99\%}$\\
MODERN&$32,300$&$32,276^{100\%}$\\
Ghani&$29,854$&$24,452^{82\%}$\\
Al-Waseet&$36,632$&$17,829^{49\%}$\\
Al-Waseet Madrasi&$7,649$&$7,384^{97\%}$\\
Thesuri$_{(7)}$&$15,236$&$12,892^{85\%}$\\
ArabicOntology\&Lexicons&$28,435$&$24,864^{87\%}$\\
ArabicWordNet&$10,929$&$9,578^{88\%}$\\
ALCSO Unified$_{(40)}$&$40,861$&$38,876^{95\%}$\\
Arab Academies$_{(16)}$&$9,675$&$7,597^{79\%}$\\
Others$_{(37)}$&$45,398$&$34,785^{77\%}$\\
Wikidata&$-$&$4665^{--}$\\
\hline
\textbf{Total}$^{110}$&$\textbf{297,608}$& $\textbf{255,528}^{84\%}$ \\ \hline
    \end{tabular}
    \vspace{-0.1cm} 
   \caption{List of lexicons mapped with \qabas so far.}
    \label{tab:lexicons}
\end{table}

\begin{table*}
    \centering
\begin{tabular}{|l|r|r|r|r|} \hline 
\textbf{Corpus}&\textbf{Tokens}& \makecell[c]{\textbf{Tokens} \\ \textbf{mapped}}&\makecell[c]{\textbf{Unique} \\ \textbf{lemmas}}& \makecell[c]{\textbf{Lemmas} \\\textbf{mapped}} \\ \hline
Arabic Treebank (MSA)&$339,710$&$282,155 ^{83\%}$&$13,078$&$12,948^{99\%}$\\ \hline
SALMA (MSA)&$34,253$&$34,253 ^{100\%}$&$3,875$&$3,875 ^{100\%}$\\ \hline
Quran (Classical)&$77,469$&$62,123 ^{80\%}$&$4,830$&$4,100^{84\%}$\\ \hline
Curras (Palestinian)&$56,169$&$56,010 ^{100\%}$&$6,033$&$5,966 ^{99\%}$\\ \hline
Baladi(Lebanese)&$9,561$&$9,493 ^{99\%}$&$2,406$&$2,365 ^{98\%}$\\ \hline
Lisan (Iraqi)&$45,881$&$40,615 ^{89\%}$&$9,306$&$7,520 ^{81\%}$\\ \hline
Lisan (Lybian)&$51,686$&$39,508 ^{76\%}$&$10,174$&$7,550 ^{74\%}$\\ \hline
Lisan (Sudanese)&$52,616$&$44,136 ^{84\%}$&$10,455$&$8,709 ^{83\%}$\\ \hline
Lisan (Yemeni)&$1,098,222$&$901,335 ^{82\%}$&$44,331$&$33,244 ^{75\%}$\\ \hline
Gummar (Emirati)&$202,329$&$182,155 ^{90\%}$&$7,590$&$6,800 ^{90\%}$\\ \hline
Nabra (Syrian)&$60,021$&$60,021 ^{100\%}$&$10,191$&$10,191 ^{100\%}$\\ \hline
Egyptian Treebank&$400,448$&$297,188 ^{74\%}$&$22,258$&$18,626^{83\%}$\\ \hline
\textbf{Total} & $\textbf{2,428,365}$& $\textbf{2,008,992}^{\textbf{83\%}}$ &$\textbf{144,527}$& $\textbf{121,894} ^{\textbf{84\%}}$ \\ \hline
    \end{tabular}
            \vspace{-0.1cm}
    \caption{List of corpora linked with \qabas so far.}
    \label{tab:corpora}
\end{table*}

\subsection{Lexicon Construction Phases}
\begin{figure*}[ht!]
    \centering
    \includegraphics[width=0.96\textwidth]{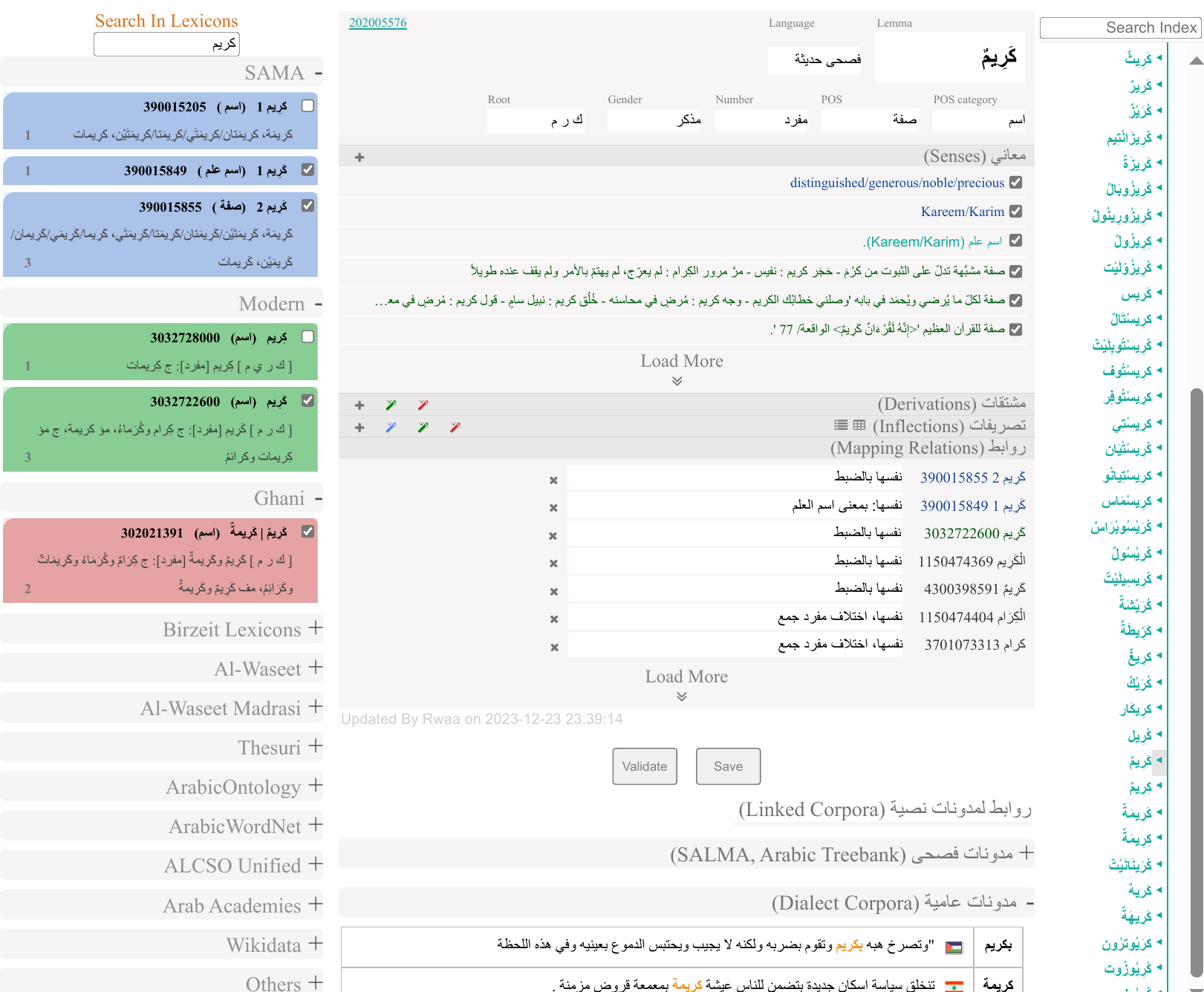}
    
    \caption{Screenshot of our web-based tool, which we developed for constructing \qabas}

\label{fig:qabas_tool}
\end{figure*}

\qabas was constructed semi-automatically over different phases, and using a web-based tool (illustrated in Figure \ref{fig:qabas_tool}). 

To bootstrap \qabas, we first adopted all lemmas from the Modern lexicon and uploaded them to the tool. Three lexicographers then reviewed and manually revised and enriched these lemmas with morphological features (described in Section \ref{sec:guidelines}) and linked them with lemmas in other lexicons. This methodology allowed the lexicographers to construct \qabas based on the information in other lexicons while linking \qabas to those lexicons at the same time (see guidelines in Section \ref{sec:guidelines}). To accelerate the linking process, we used heuristic rules to automatically discover candidate mappings for the lexicographers to verify (see Section \ref{sec:auto_mapping}).

To cover the remaining lemmas in lexicons other than Modern (i.e., that are not linked in the previous phase), we collected these lemmas and prioritized them. Higher priority is given to those lemmas that are more frequent across the $110$ lexicons and $12$ corpora. This prioritized list of candidate lemmas was uploaded to the tool, for the lexicographers to review and make the necessary edits. This approach allowed us to efficiently expand the lemma coverage of \qabas. The expansion is an ongoing and open-ended endeavor, as there is no limit to the number of lemmas that could potentially be added to \qabas. As will be discussed in section \ref{sec:evaluation}, our progress indicates that we have covered most of the lemmas in the $110$ lexicons and $12$ corpora.

Mapping \qabas with the $12$ corpora (in table \ref{tab:corpora}) was straightforward. As most of the lemmas in these corpora are SAMA lemmas, which we manually linked with \qabas, we only replaced SAMA lemmaIDs with \qabas lemmaIDs. For the non-SAMA lemmas, we selected the most frequent lemmas in the $12$ corpora and added them to \qabas manually.

\subsection{Guidelines}
\label{sec:guidelines}
Each lemma in \qabas is tagged with the following eight morphological features: (1) the $41$ POS tagset shown in Table \ref{tab:qabas_modern_sama}, (2) the gender tags \{$Masculine$, $Feminine$, $N/A$\}, (3) Number tags {\small \{$Singulare$, $Dual$, $Plural$\}}, (4) the Aspect tags {\small \{$PV$, $IV$, $CV$, $PV\_PASS$, $IV\_PASS$\}}, (5) and Person tags {\small\{$1^{st}$, $2^{nd}$, $3^{rd}$\}}. We additionally tag each lemma with its (6) root(s), (7) augmentation 
{\small\{$Augmented$, $Unaugmented$\}}, and (8) transitivity {\small\{$Transitive$, $Intransitive$\}}.

\textbf{Lemma selection and spelling}, our full list of guidelines not included in this article for space limitation but can be found online\footnote{Guidelines {\small \href{https://sina.birzeit.edu/qabas/about}{https://sina.birzeit.edu/qabas/about}}}. Our guidelines are similar to those described in the introduction of the Modern \cite{omar2008contemporary}. However, we introduced additional guidelines, such as: 
the lemma should be fully diacritized including the last letter; 
the POS of a lemma can be $Noun\_Prop$ only if all of its meanings refer to proper nouns; 
additional spellings of the same lemma are separated by "|" and ordered by frequency, such as (\Ar{تِلِيفُونٌ}|\Ar{تِلِفُونٌ}); dialectal lemmas are spelled according to the CODA rules used in Curras \cite{JHRAZ17,JHAZ14}, hence we write (\TrAr{قَزاز}) rather than (\TrAr{أزاز});
each dialect lemma is mapped with an MSA lemma, e.g.  (\TrAr{قَزاز}) and its MSA (\TrAr{زُجاج});  
a lemma is considered \small{$adjective$} if all of its meanings are either 
\scriptsize{$Active Participle$} \Ar{\tiny{اسم فاعل}}, \scriptsize{$Passive Participle$} \Ar{\tiny{اسم مفعول}},
\scriptsize{$Relative$} \Ar{\tiny{نسبة}}, 
\scriptsize{$Adjectival Propriety$} \Ar{\tiny{صفه مشبهة}}, \scriptsize{$Exaggeration$} \Ar{\tiny{صيغة مبالغة}}, or \scriptsize{$Diminutive$} \Ar{\tiny{تصغير}}  \normalsize ; among other guidelines.



\section{Lemma Linking}
\label{sec:lemma-linking}
This section presents the framework and methods we used to map between lemmas across lexicons.

\subsection{Mapping Framework}
This framework aims to enable lemmas to be interlinked through a mapping correspondence.  

\textbf{Definition $1$:} Given two lemmas $l_{1}$ and $l_{2}$, a \textit{mapping correspondence} between them is defined as:         
\begin{center}
\vspace{-0.2cm}
    <$l_{1}$, $l_{2}$, $R_i$> 
\vspace{-0.3cm}
\end{center}
Where:
\begin{itemize}
    \item $l_{1}$, $l_{2}$ are lemmas to be mapped.
    \item $R_i$ is the mapping relation between $l_{1}$ and $l_{2}$, $R_i$ \begin{math} \in \end{math} \{$R_{1}$...$R_6$\} shown in Table \ref{tab:mapping_relations}.
\end{itemize}

This mapping framework was implemented in our tool (See Figure \ref{fig:qabas_tool}) and used by our lexicographers. 
Table \ref{tab:mapping_relations} presents the count of the mapping correspondences for each relation, which are about $256K$ correspondences in total.

\begin{table}
\centering
\scriptsize 
\begin{tabular}{|wl{0.1cm} wr{1.8cm} wl{3.4cm}|| r|}
\hline
\multicolumn{3}{|c||}{\textbf{Relations}} & \multicolumn{1}{c|}{\textbf{count}} \\ \hline  
 $R_{1}$ & \tiny{\<نفسها بالضبط>} & Same Exactly & 248,882\\ \hline
 $R_{2}$ & \tiny{\<نفسها، اختلاف مفرد جمع>} & Same, Singular-Plural difference & 3,010 \\ \hline
 $R_{3}$ & \tiny{\<نفسها، اختلاف مفرد مثنى>} &  Same, Singular-Dual difference & 74 \\ \hline
 $R_{4}$ & \tiny{\<نفسها، اختلاف مذكر مؤنث>} & Same, Male-Female difference & 1,784 \\ \hline
 $R_{5}$ & \tiny{\<نفسها، اختلاف حالة إعرابية>} & Same, Case difference & 372 \\ \hline
 $R_{6}$ & \tiny{\<نفسها، بمعنى اسم العلم>} & Same, but Proper Noun & 1,918\\ \hline
 \multicolumn{3}{|c||}{\textbf{Total (mapping correspondences)}} & \multicolumn{1}{l|}{\textbf{256,040}} \\ \hline  
\end{tabular}
\vspace{-0.1cm}
\caption{The six mapping relations and their counts}
\vspace{-0.6cm}
\label{tab:mapping_relations}
\end{table}

\subsection{Automatic Mapping} 
\label{sec:auto_mapping}
To speed up the mapping process, this section proposes a set of heuristic rules to discover candidate mappings. Before presenting these rules, we discuss how Arabic word forms can be compared.

Comparing words in Arabic is not trivial. First, Arabic is diacritic-sensitive, thus we cannot compare words using equality. For example, the same lemma in one lexicon might be spelled as \TrAr{كَلمَة}\xspace and in another as \TrAr{كلَمةٌ}. Second, lexicons are not always self-consistent or follow the same guidelines in structuring or writing word forms \cite{ADJ19_report}. For example, some lexicons provide the feminine and masculine forms of a perfect verb \{\TrAr{يَكتب}, \TrAr{تَكتُب}\}, while others provide one \{\TrAr{يكتُب}\} or none \{\}. To overcome these challenges, when comparing word forms, we implemented the following definitions of \textit{compatibility} - as explained in \cite{JZAA18}.

\textbf{Definition $2$}:
Given two words $w_{1}$ and $w_{2}$, we consider them \textit{diacritic-compatible}, \textit{iff}: (1) both words have the same letters, and (2) no contradictions between the diacritics of the same, pair-wise, letters of these words.

\textbf{Definition $3$}:
Given two sets of words \textbf{$W_{1}$} and \textbf{$W_{2}$}, we consider these sets \textit{compatible}, \textit{iff} there exists a diacritic-compatible word $w$ in both sets, $w$ \begin{math} \in \end{math} \textbf{$W_{1}$} and $w$ \begin{math} \in \end{math} \textbf{$W_{2}$}, 
i.e., their intersection is not empty.

The mapping heuristic rules are:
\begin{itemize}
    \item \textbf{$h_1$:} A mapping correspondence is established between two verb lemmas if the following two conditions are true: ($i$) each lemma has a perfective form(s) {\scriptsize $PV$} and these forms are compatible, and ($ii$) if each lemma has root(s), imperfect form(s) {\scriptsize $IV$} and command form(s) {\scriptsize $CV$}, and these roots, {\scriptsize$IV$s}, and {\scriptsize $CV$s} are compatible. 

{\fontsize{9}{5}\selectfont\vspace{-0.1cm}\textbf{Example}: ($i$) $PV_1$=\{\Ar{كَتَبَ}\} and $PV_2$=\{\Ar{كَتبَ}\} which are compatible, and ($ii$) $IV_1$=\{\Ar{يكتب}, \Ar{أكتب}\} and $IV_2$=\{\Ar{يَكْتُبُ}\}, \vspace{-0.1cm} $CV_1$=\{\} and $CV_2$=\{\Ar{أكتب}\}, and $root_1$=\{\Ar{ك ت ب}\} and $root_2$=\{\}, which are all compatible.}

    \item \textbf{$h_2$:} A mapping correspondence is established between two noun lemmas if the following two conditions are true: ($i$) each lemma has a singular form(s) and these forms are compatible, and ($ii$) if each lemma has root(s), dual(s) and plural(s), and these root(s), dual(s), and plural(s) are compatible. 
\end{itemize}

With these heuristics, we were able to discover $179K$ candidate mapping correspondences. We then uploaded these mapping relations to the tool and labeled them with "Auto-mapped". Lexicographers were given these mappings to confirm and assign them one of the six relations (See the relations division at the bottom of Figure \ref{fig:qabas_tool}). Lexicographers can edit these relations and search the lexicons to include more mappings if needed.

\section{Evaluation and Discussion}
\label{sec:evaluation}
We evaluate the coverage of \qabas by comparing it with two resources: SAMA and Modern, which are well-developed resources for Arabic. SAMA is designed for morphological modeling, while Modern is a typical MSA lexicon focusing on semantics. 
Table \ref{tab:qabas_modern_sama} shows that \qabas's coverage is almost double of Modern and is $40\%$ larger than SAMA. Table \ref{tab:lexicons} also shows that \qabas contains all Modern lemmas and $99\%$ of SAMA lemmas. We did not add the $1\%$ as we found them to be typos or with redundant spellings.
Another critical issue in SAMA is that it treats each proper noun as a separate lemma (e.g., \TrAr{كَرِيم1}\xspace as a proper noun and \TrAr{كَرِيم2}\xspace as adjective). 
We believe that this is problematic because most Arabic words can be used as proper nouns \cite{JKG22}. Proper nouns in \qabas are considered as such only if all meanings denote proper nouns. Thus, the lemma \TrAr{كَرِيم}\xspace would be tagged with an adjective, and one of its meanings is a proper noun. Hence, most of the $5,540$ proper nouns in SAMA are merged and mapped with \qabas lemmas through the $R_6$ relations.

\textbf{An Inter-Annotator Agreement (IAA)} evaluation was conducted to evaluate the lemma mappings. We randomly selected $2850$ lemmas ($5\%$ of \qabas) and asked each of the three lexicographers ($A_1$, $A_2$, $A_3$) to map them. The IAAs using the Kappa coefficient $\kappa$ are: $A_1$-$A_2$ is $85\%$, $A_2$-$A_3$ is $88\%$, and $A_1$-$A_3$ is $86\%$, which are "almost perfect" \citep{viera2005understanding}.

\begin{table}[ht!]
\centering
\scriptsize
\setlength\tabcolsep{4pt}
\begin{tabular}{|l|l|r|r|r|}
\hline
\textbf{\makecell{POS\\category}} &
  \textbf{POS} &
  \multicolumn{1}{l|}{\textbf{Modern}} &
  \multicolumn{1}{l|}{\textbf{SAMA}} &
  \multicolumn{1}{l|}{\textbf{\qabas}} \\ \hline
\multirow{11}{*}{\textbf{\rotatebox[origin=c]{90}{\textbf{Nominal}}}} & NOUN   \tiny{\<اسم>}         & 21,456          & 19,705          & 29,053          \\ \cline{2-5} 
                                   & NOUN\_PROP  \tiny{\<اسم علم>}   &               & 5,540           & 4,319           \\ \cline{2-5} 
                                   & ADJ     \tiny{\<صفة>}       &               & 5,500           & 11,067          \\ \cline{2-5} 
                                   & ADJ\_COMP \tiny{\<صفة مقارنة>}     &               & 204            & 295            \\ \cline{2-5} 
                                   & ADJ\_NUM   \tiny{\<صفة عدد>}    &               & 12             & 12             \\ \cline{2-5} 
                                   & NOUN\_NUM  \tiny{\<اسم عدد>}    &               & 33             & 44             \\ \cline{2-5} 
                                   & NOUN\_QUANT  \tiny{\<اسم كم>}  &               & 23             & 19             \\ \cline{2-5} 
                                   & DIGIT      \tiny{\<عدد>}    &               &               &  10             \\ \cline{2-5} 
                                   & NOUN\_VOICE \tiny{\<صوت>}   &               &               & 16             \\ \cline{2-5} 
                                   & ABBREV     \tiny{\<حرف اختصار>}    &               & 60             & 106             \\ \cline{2-5} 
                                   & \textbf{Total} & \textbf{21,456} & \textbf{31,077} & \textbf{44,941} \\ \hline
\multirow{6}{*}{\textbf{\textbf{\rotatebox[origin=c]{90}{\textbf{Verb}}}}}     & PV  \tiny{\<ماضي>}            & 10,475              & 8,133           & 12,679          \\ \cline{2-5} 
                                   & IV   \tiny{\<مضارع>}          &               & 990            & 9              \\ \cline{2-5} 
                                   & CV  \tiny{\<أمر>}            &               & 16             & 6              \\ \cline{2-5} 
                                   & PV\_PASS   \tiny{\<ماضي مجهول>}     &               & 32             & 63             \\ \cline{2-5} 
                                   & IV\_PASS   \tiny{\<مضارع مجهول>}     &               & 78             &               \\ \cline{2-5} 
                                   & \textbf{Total} & \textbf{10,475} & \textbf{9,249}  & \textbf{12,757} \\ \hline
\textbf{\rotatebox[origin=c]{90}{\textbf{\makecell{Functional \\words}}}} &
  \tiny{\makecell[l]{
  PRON, DEM\_PRON, EMOJI \\
  REL\_PRON, REL\_ADV,\\ 
  ADV, INTERROG\_PART, \\ 
  INTERROG\_ADV,PREP,CONJ,\\ 
  INTERROG\_PRON, PART\\ 
  RESTRIC\_PART,PUNC,INTERJ,\\
  FOCUS\_PART, DET, VERB\\ 
  VOC\_PART, PROG\_PART, \\
  SUB\_CONJ, VERB\_PART,\\
  FUT\_PART,EXCLAM\_PRON\\
  PSEUDO\_VERB,NEG\_PART \\}}
  &
  369 &
  313 &
  473 \\ \hline \hline
  
&\textbf{Total}&\textbf{32,300}&\textbf{40,639}&\textbf{58,171}  \\ \hline
\end{tabular}
\caption{Coverage Evaluation of \qabas, per POS}
\vspace{-0.5cm}
\label{tab:qabas_modern_sama}
\end{table}

\section{Conclusion}
\label{sec:conclusion}
We presented \qabas, a novel and open-source Arabic lexicon linked with $110$ lexicons and $12$ morphologically annotated corpora. Additionally, the $256k$ mappings correspondences between \qabas and each of the $110$ lexicons can be also downloaded from \href{https://sina.birzeit.edu/qabas/about}{Qabas Page}. As such, \qabas is a large lexicographic data graph, linking existing Arabic lexicons and annotated corpora.

\subsection{Limitations and Future Work}
One of the major challenges faced during the construction of \qabas was convincing the owners of the lexicons to publish their lexicons as open-source.
 While we agreed with the owners of the lexicons to only publish the mapping links between \qabas and their lexicons, we hope that our work will encourage others to publish their lexicons as open-source in the future. 
 Adding dialect lemmas to \qabas is another challenge. 
 Since our three lexicographers are familiar with Levantine dialects, adding lemmas from other dialects requires knowledge of these dialects. 
 \qabas is currently limited to the frequently used dialect lemmas or those that are known to most Arabs. 
 We plan to recruit more lexicographers from other dialects to extend \qabas. 
 Last but not least, we plan to represent \qabas and publish the mapping correspondences using the W3C RDF Lemon model.

\subsection{Ethical and copyright Considerations}
\label{sec:ethical_considerations}
We obtained permission to use the lexicons and corpora listed in this article, and since our lexicon will be open-source, we will not share any copyrighted data. We will share: (1) \qabas itself (all lemmas and their full morphological features), and (2) the mapping links (i.e., correspondences) between \qabas and the other external resources. Obtaining licenses for these external resources is the responsibility of the users.

\section*{Acknowledgment}
\label{sec:ack}
We would like to thank the main lexicographers who contributed to this project, especially Shimaa Hamayel, Hiba Zayed, and Rwaa Assi; as well as Diyam Akra, Sanad Malaysha, Sondus Hamad, Asmaa Motan, Yaqout Abu Allia, Nour Dana, who also contributed to various lexicographic and technical aspects.


\section{References}
\label{sec:references}

\bibliographystyle{lrec-coling2024-natbib}
\bibliography{lrec-coling2024-example,MyReferences}
\bibliographystylelanguageresource{lrec-coling2024-natbib}
\bibliographylanguageresource{languageresource}

\end{document}